\documentclass[sn-basic]{sn-jnl}

\usepackage{graphicx}%
\usepackage{multirow}%
\usepackage{amsmath,amssymb,amsfonts}%
\usepackage{amsthm}%
\usepackage{mathrsfs}%
\usepackage[title]{appendix}%
\usepackage{xcolor}%
\usepackage{textcomp}%
\usepackage{manyfoot}%
\usepackage{booktabs}%
\usepackage{algorithm}%
\usepackage{algorithmicx}%
\usepackage{algpseudocode}%
\usepackage{listings}%
\usepackage{makecell}
\usepackage{varwidth}
\usepackage{orcidlink}
\usepackage{tabularx}
\usepackage{pdflscape}

\begin{document}
\renewcommand*{\thefootnote}{\alph{footnote}}

\title[An Overview Of Temporal Commonsense Reasoning and Acquisition]{An Overview Of Temporal Commonsense Reasoning and Acquisition}


\author*[1]{\fnm{Georg} \sur{Wenzel} \orcidlink{0009-0006-7867-8876}}\email{georg.wenzel@student.uibk.ac.at}

\author[1]{\fnm{Adam} \sur{Jatowt} \orcidlink{0000-0001-7235-0665}}\email{adam.jatowt@uibk.ac.at}

\affil[1]{\orgdiv{Department of Computer Science}, \orgname{University of Innsbruck}, \orgaddress{\street{Technikerstraße 21a}, \city{Innsbruck}, \postcode{6020}, \state{Tyrol}, \country{Austria}}}

\abstract{Temporal commonsense reasoning refers to the ability to understand the typical temporal context of phrases, actions, and events, and use it to reason over problems requiring such knowledge. This trait is essential in temporal natural language processing tasks, with possible applications such as timeline summarization, temporal question answering, and temporal natural language inference. Recent research on the performance of large language models suggests that, although they are adept at generating syntactically correct sentences and solving classification tasks, they often take shortcuts in their reasoning and fall prey to simple linguistic traps. This article provides an overview of research in the domain of temporal commonsense reasoning, particularly focusing on enhancing language model performance through a variety of augmentations and their evaluation across a growing number of datasets. However, these augmented models still struggle to approach human performance on reasoning tasks over temporal common sense properties, such as the typical occurrence times, orderings, or durations of events. We further emphasize the need for careful interpretation of research to guard against overpromising evaluation results in light of the shallow reasoning present in transformers. This can be achieved by appropriately preparing datasets and suitable evaluation metrics.}



\maketitle

\section{Introduction}
\label{sec:introduction}
Humans generally perform well in interpreting implicit information in text and speech by leveraging \emph{commonsense reasoning}. This ability is reflected in the way we communicate. For example, when we read the phrase ``I couldn't get out of bed this morning.'', we generally assume that this refers to a state of mind and not a physical inability to get out of bed. When we read ``He had butterflies in his stomach.'', we understand this as a figure of speech for an anxious or nervous feeling. Rather than specifying the literal meaning, we rely on the recipient's implicit prior understanding of certain concepts and expressions in our language.

Commonsense reasoning can manifest in different forms. Datasets such as \textsc{CIDER} \citep{cider}, \textsc{Cosmos QA} \citep{cosmosqa}, \textsc{GLUCOSE} \citep{glucose}, and \textsc{COM2SENSE} \citep{comsense} aim to serve as benchmarks to better understand the commonsense reasoning capabilities of current state-of-the-art machine learning models. In the process, these capabilities are often grouped into taxonomies, composed of categories such as physical common sense, social common sense, motivations, reactions, causality, and several others. Furthermore, collecting commonsense knowledge can be a primary goal for some knowledge bases, such as the \textsc{ConceptNet} \citep{conceptnet} and \textsc{ATOMIC}$^{20}_{20}$ \citep{atomic} \emph{knowledge graphs} (KGs), which have the goal of both bolstering the general reasoning capabilities of \emph{language models} (LMs) and training them to be able to express their implicit knowledge directly for evaluation purposes.

Historically, building machine learning systems with commonsense reasoning was a problem that was relatively difficult to tackle. One of the reasons for the first AI winter, a period of reduced funding and interest in artificial intelligence, was the lack of algorithmic problem-solving approaches, with many developers instead attempting to build systems that ``think humanly'' \citep{aiwinter}. However, due to advances in computing power and neural models, these approaches have seemingly become possible in many \emph{natural language processing}~(NLP) tasks \citep{gpt-2}. A driving force behind this change is the use of transformer models \citep{transformer} and the LMs they enable, such as \textsc{BERT} \citep{bert} and \textsc{GPT} \citep{gpt}. 

This article focuses on \emph{temporal commonsense}~(TCS) reasoning. TCS encompasses a variety of traits. For example, given the pair of sentences ``Mary went to the hospital. She broke her leg.'', the likely sequence of events is that Mary first broke her leg and then went to the hospital, despite this not being explicitly expressed in the text. Understanding event durations is another such property. We intuitively know that going on a walk takes less time than going on vacation, even though the structure of both phrases is very similar. 

Although the specific notion of TCS is relatively new, many of its applications are not. In this survey, we first provide some background on the field of temporal reasoning, where some tasks, such as event relation extraction, which directly relate to proposed TCS dimensions, have already been explored since the early 2000s \citep{time-ml,timebank,tempeval-2007}.

In addition to the apparent benefit of incorporating TCS reasoning into such tasks, time-aware LMs for downstream NLP tasks are also becoming increasingly popular. Recently, models such as \textsc{TempoBERT} \citep{tempobert} and \textsc{BiTimeBERT} \citep{timebert} have been proposed, which aim to temporalize the embeddings provided by LMs such as \textsc{BERT} via the document creation time or explicit temporal expressions in the training corpus. Another approach is to temporalize the attention mechanism of the transformer itself \citep{tempattention}. Generally, these approaches are evaluated in domains such as semantic change detection or document dating, where the use of explicit timestamps may not only be encouraged by the available datasets, but may even be required to perform the task in the first place. These models often outperform previous non-transformer-based state-of-the-art solutions in their respective domains. 

Similarly, LMs with TCS may achieve higher performance in domains where explicit temporal expressions or document dates are not as widely available. \citet{stack-sentence-order} utilize the \textsc{COMET} transformer model \citep{atomic}, which was trained on the \textsc{ATOMIC}$^{20}_{20}$ KG, to incorporate commonsense knowledge such as ``I called 911 to report the accident.'' occurring before ``The police soon arrived.'' into a sentence ordering task, achieving state-of-the-art results on several datasets. \citet{zhang-audio} use temporal knowledge embedded in the \textsc{ASER} \citep{aser} KG to enrich an audio tagging ontology. LMs with more precise world models, including an understanding of TCS properties such as typical event orderings or durations, could also be helpful in tasks such as timeline summarization \citep{tls-covid19}, sequencing \citep{sequencing} or question answering \citep{temporal-qa}. Tasks such as timeline summarization and question answering, while currently often based on and evaluated against document collections with explicit document creation times, can also rely heavily on the contextual understanding of a user's query and the temporal interaction between documents. 

In the remainder of this survey, we mainly focus on recently proposed benchmark datasets and LMs incorporating TCS. From this research, we can draw various conclusions for future work, analyse the currently best-performing methods, and identify research gaps. The rest of this article is structured as follows. Section~\ref{sec:collection} describes our method to collect relevant literature for this survey and lists related work. Section~\ref{sec:background} provides background knowledge regarding the field of temporal reasoning. Section~\ref{sec:pretransformer} illustrates the shift from structural, rule-based reasoning to a more data-driven approach on several NLP tasks, the different types of TCS knowledge, and pre-transformer approaches. Section~\ref{sec:sota} lists recent benchmark datasets and examines proposed ways to improve them. Section~\ref{sec:approaches} gives an overview of how researchers have been attempting to improve performance on TCS tasks in recent years. In Section~\ref{sec:discussion}, we propose possible avenues for future work and discuss the current state of the art. Finally, Section~\ref{sec:summary} summarizes the content of the article and provides an outlook for future research.

\section{Survey Scope and Related Work}
\label{sec:collection}

In this section, we illustrate the scope of this literature review by placing the field of TCS reasoning in its surrounding context within the NLP landscape. This allows us to clearly define which type of research will be included in the survey. Notable differences from recent related literature are also highlighted. 

\subsection{Survey Scope}
\begin{figure}[htbp]
    \centering
    \includegraphics[width=119mm]{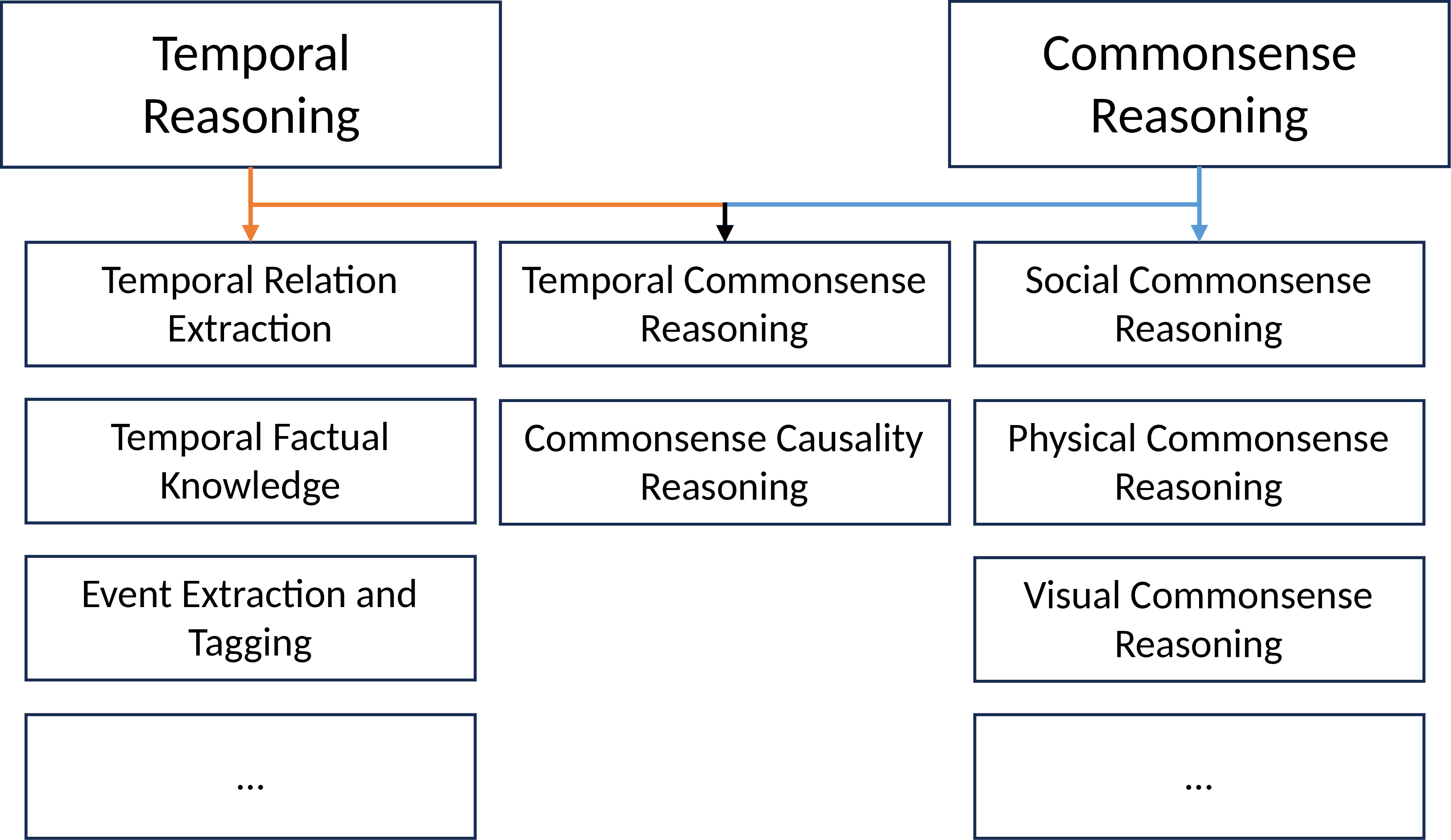}
    \caption{Temporal reasoning and commonsense reasoning both encapsulate TCS reasoning, but also contain many other tasks}
    \label{fig:tcs}
\end{figure}

The field of TCS reasoning is semantically embedded within both commonsense and temporal reasoning. Figure \ref{fig:tcs} shows some example tasks from both domains. In this survey, we focus specifically on datasets and models for TCS reasoning, in contrast to related fields, which we will discuss in this section.

\subsubsection{Commonsense Reasoning}
As noted in Section \ref{sec:introduction}, many datasets exist already to benchmark different types of common sense. Typically, such datasets focus on several dimensions of common sense. Consequently, TCS was considered just one of several categories or even completely overlooked. We choose not to survey such datasets, as more recent research provides several datasets specifically to benchmark TCS reasoning. Additionally, it is likely that many models specifically developed to reason over temporal properties would not perform well on other types of commonsense reasoning. 

\subsubsection{Temporal Reasoning}
This category encompasses a wide range of research. In this survey, we do not focus on purely algorithmic approaches for temporal reasoning, such as dependency tree parsing or logical propositions. We also differentiate between \emph{temporal factual knowledge}, where an LM is evaluated on its knowledge of the temporal scope of certain facts \citep{templama}, and \emph{TCS knowledge}, which is centred around an implicit understanding of common temporal attributes. For example, knowing that a presidential term has a duration of years rather than minutes is TCS. However, knowledge of the identity of the President of the United States in 2009 is temporal factual knowledge. One possible ambiguity emerges when temporal factual knowledge tasks, such as temporal slot filling, are tackled using common sense that is not inherently temporal, such as knowledge of certain ``world invariants'' \citep{temporalslotfilling-commonsense,temporalfactextraction-commonsense}. However, to keep the scope of the survey reasonable, we do not explore such approaches, as they technically do not leverage TCS reasoning.

\subsubsection{Commonsense Causality Reasoning}
Commonsense causality reasoning is perhaps the most closely related field to TCS reasoning. Like TCS reasoning, it finds its roots in both temporal and commonsense reasoning. Naturally, temporal awareness is almost certainly required to reason about causality \citep{rock}. Conversely, causality can greatly inform certain TCS dimensions, such as event ordering. However, causality and the properties proposed in TCS reasoning are ultimately different. In line with our goal of keeping the scope of the survey reasonable, we thus do not study such approaches.

\subsection{Research Goals}
In this survey, we aim to provide a broad overview of the field of TCS reasoning. This includes scoping TCS, as well as identifying and collecting relevant datasets, LM structures, evaluation metrics, and state-of-the-art results. 

Our first major objective is to provide a full overview of datasets specifically developed to evaluate certain dimensions of TCS, as well as survey said datasets for common evaluation metrics, findings, and possible identified methods to improve the robustness of both the collection process for new datasets and the reporting process for existing ones.

We find that current research into LMs with TCS mainly revolves around data engineering of input- and output structures of the transformer architecture. Consequently, our second major goal is to summarize and categorize the attempted augmentations as well as their perceived effectiveness, and to locate possible avenues for future work.

\subsection{Literature Collection}
The literature collection for this survey was conducted as follows. For the primary collection of state-of-the-art TCS models and datasets, \emph{Google Scholar}, \emph{Semantic Scholar}, and \emph{dblp} were queried using the search string ``temporal commonsense''. Additionally, we restricted the field of study to ``computer science'' on the Semantic Scholar platform.

The transformer architecture and subsequent models significantly improved the state-of-the-art performance in many NLP tasks \citep{gpt-2}. This survey will show that transformers form the basis of nearly all state-of-the-art models in TCS reasoning. Therefore, we only considered research from 2018 to 2023, in line with the 2017 release of the ``Attention is all you need'' paper \citep{transformer} and the subsequent 2018 release of the \textsc{BERT} and \textsc{GPT} models. 

We evaluated 21 papers from dblp (the full set of results) and the top 50 results from Google Scholar and Semantic Scholar, discarding research that was not written in English or did not mention TCS in the abstract. We then performed one iteration of backward snowballing from the result set to identify previous work. However, except for the \textsc{ROCStories} dataset, most previous work could not clearly be described as aiming to acquire or measure TCS understanding.

First, we provide a brief background on temporal reasoning to ground the origin of tasks and dimensions proposed within TCS reasoning. Our main objective is then the collection of modern benchmark datasets for TCS reasoning, as well as proposed models evaluated on said datasets. A summary of the most important literature is shown in Table \ref{tab:overview}. Note that we prioritize work that matches the TCS domain in this table, and out-of-domain resources (e.g., out-of-domain datasets used for evaluation) may not be fully listed in Table \ref{tab:overview}. On top of the articles highlighted in this table, we cite important work from the previously discussed related domains to highlight findings that can be applied to TCS reasoning research in the future.

\begin{table}[!h!t]
    \footnotesize
    \tabcolsep=0.11cm
    \centering
    \caption{A summary of the most important literature to our survey. The main contribution to the field is highlighted in bold for each paper. Augmentations are introduced in the main sections of this paper. (EXT - External Knowledge; ENC - Data Encoding; LSR - Logical or Symbolic Reasoning; ADV - Adversarial Learning; WS - Weak Supervision; ENS - Ensemble)}
    \begin{tabular}{llllll}
    \toprule
    Paper & Dataset(s) & Model & Aug. & Task & Metrics \\
    \midrule
    \citet{rocstories} & \textbf{\textsc{ROCStories}} & DSSM &  & Story Completion & Acc \\ \hline
    \citet{matres} & \textbf{\textsc{MATRES}} & Perceptron &  & Order Prediction & P, R, F1 \\ \hline
    \citet{content-expiry-date} & \textbf{\textsc{Alm2019}} & SVC & \textbf{WS} & \textbf{TV Duration} & F1 \\ \hline
    \citet{mctaco} & \textbf{\textsc{McTaco}} & \textsc{BERT} & \textbf{ENC} & \textbf{Cloze QA} & F1, EM \\ \hline
    \citet{commonsense-storygen} & \textsc{ROCStories} & \textsc{GPT-2} & \textbf{EXT} & Story Completion & \makecell[l]{Human, PPL\\BLEU, Cov.,\\Rep., distinct-4} \\ \hline
    \citet{torque} & \textbf{\textsc{TORQUE}} & \textsc{RoBERTa} &  & Order Prediction & \makecell[l]{F1, EM, \\Consistency} \\ \hline
    \citet{alice} & \textsc{McTaco} & \textsc{RobERTa} & \textbf{ADV} & Cloze QA & F1, EM \\ \hline
    \citet{wikihow} & \textbf{\textsc{WikiHow}} & \textsc{RoBERTa} &  & Step Ordering & Acc \\ \hline
    \citet{Vashishtha2020} & \textbf{\textsc{Vash2020}} & \textsc{RoBERTa} &  & \makecell[l]{Duration Inference\\Order Inference} & Acc \\ \hline
    \citet{timeawarept} & \textsc{McTaco} & \textsc{BERT} & \makecell[l]{\textbf{WS}\\\textbf{ENC}} & Duration Prediction & F1, EM \\ \hline
    \citet{contrastsets} & \textsc{McTaco} & \textsc{RoBERTa} & & Cloze QA & \makecell[l]{\textbf{Contrast EM}, \\\textbf{Consistency}} \\ \hline
    \citet{selfexplain} & \makecell[l]{\textsc{SST-5}\\\textsc{SNLI}} & \textsc{RoBERTa} & \textbf{ENC} & NLI & Acc, F1, BLEU \\ \hline
    \citet{taco-lm} & \textsc{McTaco} & \textsc{BERT} & \makecell[l]{\textbf{EXT}\\\textbf{WS}} & Cloze QA & \makecell[l]{F1, EM} \\ \hline
    \citet{timedial} & \textbf{\textsc{TIMEDIAL}} & \textsc{BERT} &  & Cloze Completion & 2-best Acc \\ \hline
    \citet{stack-sentence-order} & \textsc{ROCStories} & \textsc{BART} & \textbf{EXT} & Sentence Ordering & Kendall's $\tau$, Acc\\ \hline
    \citet{alicepp} & \makecell[l]{\textsc{MATRES}\\\textsc{McTaco}} & \textsc{RoBERTa} & \textbf{ADV} & \makecell[l]{Order Prediction\\Cloze QA} & Acc, F1, EM \\ \hline
    \citet{zhang-audio} & \makecell[l]{\textsc{SONYC}\\\textsc{AudioSet}} & D-GCN & \textbf{EXT} & Audio Tagging & mAP, mAUC \\ \hline
    \citet{ensemble-bert-tcs} & \textsc{McTaco} & \textsc{BERT} & \makecell[l]{\textbf{ENS}\\\textbf{EXT}} & Cloze QA & F1, EM \\ \hline
    \citet{tracie} & \textbf{\textsc{TRACIE}} & \textsc{T5} & \textbf{LSR} & Order Inference & Acc \\ \hline
    \citet{tempwikibio} & \textsc{TempWikiBio} & \textsc{BART} & \textbf{ENC} & Text Completion & \makecell[l]{Human, BLEU\\METEOR,\\BERTScore} \\ \hline
    \citet{tempattention} & \textsc{SemEval} & \textsc{BERT} & \textbf{ENC} & Semantic Change & \makecell[l]{Pearson's $r$,\\Spearman's $\rho$} \\ \hline
    \citet{tempobert} & \makecell[l]{\textsc{LiverpoolFC}\\\textsc{SemEval},\textsc{NYT}} & \textsc{BERT} & \textbf{ENC} & Semantic Change & \makecell[l]{Pearson's $r$,\\Spearman's $\rho$} \\ \hline
    \citet{timebert} & \makecell[l]{\textsc{EventTime}\\\textsc{WOTD}\\\textsc{NYT},\textsc{TDA}} & \textsc{BERT} & \textbf{ENC} & \makecell[l]{Occurrence Time\\Document Dating} & \makecell[l]{Acc, MAE,\\F1, EM} \\ \hline
    \citet{eventoptimaltransport} & \textsc{McTaco} & \textsc{BART} & \textbf{ENC} & Cloze QA & Acc, F1 \\ \hline
    \citet{sleer} & \textsc{McTaco} & \textsc{BERT} & \textbf{LSR} & Cloze QA & F1, EM \\ \hline
    \citet{cocolm} & \makecell[l]{\textsc{ROCStories}\\\textsc{MATRES}} & \textsc{BERT} & \textbf{EXT} & \makecell[l]{Order Prediction\\Story Completion} & Acc, F1 \\ \hline
    \citet{tnli} & \textbf{\textsc{TNLI}} & \textsc{RoBERTa} & \textbf{EXT} & \textbf{TV Inference} & Acc \\ \hline
    \citet{ssl-induction} & \textsc{TIMEDIAL} & \textsc{RoBERTa} & \textbf{LSR} & Cloze Completion & 2-best Acc \\ \hline
    \citet{cta} & \textbf{\textsc{CoTAK}} & \textsc{BERT} & & \makecell[l]{Duration Prediction\\TV Duration} & Acc \\
    \botrule
    \end{tabular}
    \label{tab:overview}
\end{table}

\begin{figure}[htbp]
    \centering
    \includegraphics[width=119mm]{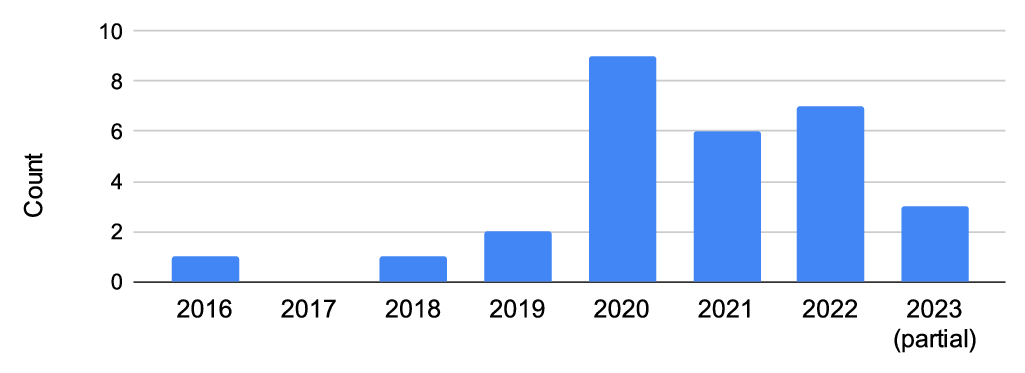}
    \caption{The distribution of publication dates in the surveyed core literature}
    \label{fig:pubdate}
\end{figure}

We further break down the information in Table \ref{tab:overview} by publication date and contribution type. Figure \ref{fig:pubdate} shows the distribution of publication dates in the core literature. Since 2020, there appears to be a constant stream of research into TCS reasoning, and this trend does not seem to fade in 2023. Based on the publication dates, it is likely that the transformer architecture is a strong contributing factor to the large number of new publications in this field.

\begin{figure}[htbp]
    \centering
    \includegraphics[width=119mm]{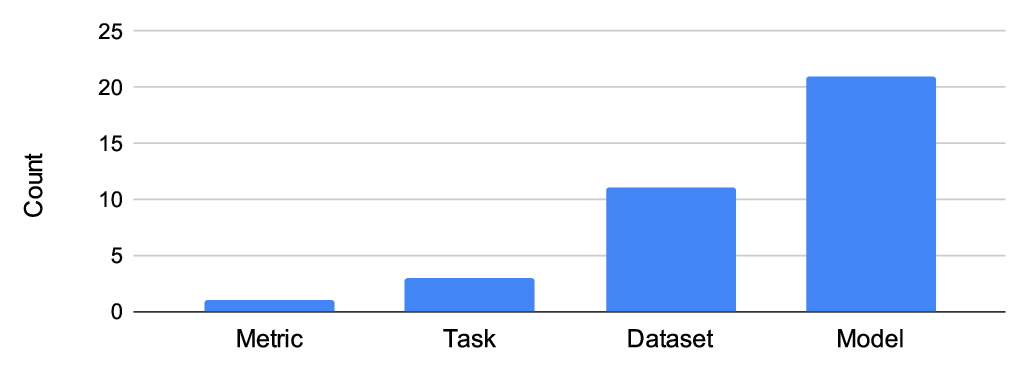}
    \caption{The distribution of contribution types in the surveyed core literature}
    \label{fig:pubtype}
\end{figure}

Figure \ref{fig:pubtype} shows the distribution of the main contribution type in the surveyed core literature. Specifically, these categories correspond to the highlighted text within Table \ref{tab:overview}. They signify whether an article proposed a new model (or augmentation) for an existing dataset, a new dataset, a new task or taxonomy, or an evaluation metric. One article may have multiple such contributions. The chart shows that there is a strong focus on models and model augmentations. This is not necessarily bad, as it shows a community effort to improve performance on well-defined problems, but a lack of focus on more rigid evaluation metrics may contribute to suboptimal evaluation practices on many of the proposed model augmentations, which will be discussed in the main sections of the survey. 

\subsection{Related Work}
Several recent surveys are studying commonsense knowledge embedded in LMs and how it could be improved \citep{commonsense-survey-2,commonsense-survey-1,multimodal-survey,knowledge-survey}. However, such surveys tend to only consider TCS as one of several possible domains of common sense, if at all, and do not provide a full spectrum of recent research. 

\citet{benchmark-survey} provides a comprehensive survey of benchmark datasets for different categories of commonsense reasoning, including temporal. However, their survey aimed to qualitatively analyse a large variety of common sense benchmark datasets to detect potential flaws and propose improvements. As there is no categorization by the type of common sense required and no further in-depth comparative discussion of results within the domain of TCS research, this survey does not provide a clear overview of the current state of the art.

\citet{pitfalls} showcase shallow reasoning behaviours in transformer models on different tasks. While not all proposed behaviours are related to commonsense reasoning, some examples from the TCS domain, also discussed in this survey, are mentioned. Furthermore, some general behaviours found in LMs, such as the possibility of mispriming and a lack of understanding for negated phrases, have substantial implications for tasks posed in the TCS domain and should be considered when training models on temporal data \citep{timedial}. 

\citet{kg-survey} survey KGs, specifically mentioning them as a possible way to empower commonsense reasoning in knowledge-aware models. However, many alternative approaches can be chosen to encode additional temporal information in LMs, which are not discussed in this survey.

Of course, surveys can also be found on downstream tasks related to time, such as temporal information retrieval \citep{tir-survey} and temporal information extraction \citep{tieval}. As mentioned in Section \ref{sec:introduction}, it stands to reason that models incorporating TCS could help solve such tasks. However, because of the relative novelty of the topic, they are not often discussed.

Compared to previous work, our survey focuses specifically on the growing field of TCS reasoning. We draw parallels to long-standing temporal reasoning tasks and highlight the technologies that enabled the current state-of-the-art performance. We then survey datasets explicitly created to benchmark the TCS understanding of machine learning models, as well as models evaluated on those datasets. Finally, from both types of surveyed research and related work, we categorize the current state-of-the-art solution space and propose improvements for both datasets and models in future work.

\section{Temporal Reasoning}
\label{sec:background}
We first briefly summarize the topic of \emph{temporal reasoning}. One major evolution in temporal reasoning tasks over the years is a shift from a more syntactic, rule-based problem-solving approach to a more semantic, data-driven one. Specifically, in this paper, we consider as \emph{syntactic} approaches algorithms or models that focus on the structural or grammatical aspects of language. For example, syntactic parse trees are used to represent and parse information from the grammatical structure of sentences. Rule-based classifications, which rely on predefined rules for identifying features, are another form of syntactic analysis we consider. On the other hand, \emph{semantic} approaches describe algorithms or models that are concerned with the meaning and interpretation of language. This includes models based on word embeddings, which are typically learned through various machine learning tasks incorporating the usual local context of words, as well as other data-driven methods that capture nuances and relationships between words and phrases implicitly through patterns in the training data.

Temporal reasoning consists of ``formalizing the notion of time and providing means to represent and reason about the temporal aspects of knowledge'' \citep{temporal-reasoning}. Much early temporal reasoning research in NLP can be linked back to \citet{allen} documenting an algebra for storing and updating temporal relations between events in the form of intervals, which were connected using a set of 13 different relations such as \emph{during}, \emph{before}, \emph{after} or \emph{overlaps}. This algebra stood out from previous work in that it did not require precise timestamps or orderings to be known and could be used to express facts such as ``event A happens before or after event B'', similar to how temporal facts can be expressed in natural language, without explicit timestamps and without the strict requirement of an ordered notion of time between events.

In text, the technology development program TIDES first led to annotation guidelines for explicit temporal expressions using the TIMEX standard \citep{tides}. The 2002 TERQAS workshop then led to the creation of the specification language \emph{TimeML} \citep{time-ml} for connecting temporal expressions to specific events in natural language. TimeML effectively converted many of the constraints previously defined by Allen into tags for natural language. \citet{time-ml} define four fundamental problems in event-temporal identification.

\begin{enumerate}
    \item Timestamping of events.
    \item Ordering events with respect to one another.
    \item Reasoning with underspecified temporal expressions (such as ``last week'').
    \item Reasoning about the persistence of events.
\end{enumerate}

TimeML defines tags such as temporal links between events and definitions for intensional temporal expressions, embeddings, and signal words. Therefore, it was suitable for annotating events and their temporal dimensions in textual content. It was subsequently used for the creation of \emph{TimeBank} \citep{timebank}, a text corpus of 300 documents from various news-related sources, manually annotated using TimeML tags. Over the following years, TimeBank was further refined to ensure that it could be used as a gold standard for temporal relation extraction \citep{timebank1-2}. 

TimeML and TimeBank proved to be crucial resources for benchmarking temporal reasoning in the following years. A notable resource that promoted the development of TimeML as an annotation language was the \emph{TempEval} tasks proposed in several SemEval workshops between 2007 and 2013 \citep{tempeval-2007,tempeval-2010,tempeval3}. Over the years, the task definitions in \emph{TempEval} were adapted to gradually require an increasing reasoning scope in recognizing, extracting, and tagging temporal expressions and events from free-form text. In this sense, the development of these challenges and the corresponding solution space highlight a potential origin of the idea of ``temporal common sense''. 

In 2007's TempEval challenge \citep{tempeval-2007}, participants were required to extract and provide simplified TimeML annotations for already supplied events and temporal links. Rule-based systems and syntactic analysis (such as dependency tree parsing or syntactic tree generation) were used to solve the task. 

In contrast, 2010's TempEval-2 tasks \citep{tempeval-2010} extended the initial TempEval task set with the automatic recognition of events and time expressions. In contrast to previous tasks regarding the extraction of temporal expressions, where rule-based information extraction systems such as the Edinburgh IE system \citep{tempeval2-edinburgh} and HeidelTime \citep{tempeval2-heideltime} dominated, a conditional random field~(CRF) model provided the best F1 score on event extraction \citep{tempeval2-tipsem}. The authors of this CRF model show that the inclusion of semantic features, such as semantic role labels or other lexical semantics such as WordNet ontology classes, can improve the model's ability to generalize and lead to a higher recall as a result. The success of this approach already showed a movement towards data-driven methods and the use of latent semantic information for classification purposes rather than purely syntactic parsing.

This trend continued in 2013, where the TempEval-3 task set \citep{tempeval3} included an end-to-end task requiring the systems to fully extract events and their temporal links from scratch and tag all extracted data with appropriate properties. The dataset used for the previous challenges was expanded with a new platinum test set containing previously unseen text with expert annotations as well as an automatically annotated silver set, using an ensemble of best-performing methods from the previous TempEval challenge. This extended dataset effectively allowed teams to leverage precomputed weak supervision. Again, while rule-based systems dominated on pure normalization of time expressions, machine-learning-based systems performed much better on the event extraction task, with all high-performing systems using some form of machine learning, usually in the form of probability classifiers such as MaxEntropy, CRF, or support vector machines~(SVM). The automatically annotated silver data and semantic features, such as WordNet synsets and semantic role labels, also proved very helpful in solving this challenge. 

This difference in best-performing solutions raises the question of what distinguishes a task like temporal expression normalization from more event-centric tasks. The following example of events in text from the TimeML specification \citep{time-ml, timeml-guidelines}, with events highlighted in bold, illustrates why it may be difficult for a rule-based system to perform event extraction.

\begin{samepage}
    \noindent\rule[1pt]{0.99\textwidth}{1pt}

    \nopagebreak
    \noindent He \textbf{[kicked]} the ball, and it \textbf{[rose]} into the air.
    
    \nopagebreak
    \noindent The \textbf{[rains]} \textbf{[caused]} the \textbf{[flooding]}.
    
    \nopagebreak
    \noindent John \textbf{[caused]} the \textbf{[fire]}.
    
    \nopagebreak
    \noindent All 75 people \textbf{[on board]} the Aeroflot Airbus \textbf{[died]}.

    \nopagebreak
    \noindent\rule[3pt]{0.99\textwidth}{1pt}
\end{samepage}

According to the TimeML annotation guidelines, events ``cover situations that happen or occur. […] We also consider as events those predicates describing states or circumstances in which something obtains or holds true''. Compared to the limited number of possible explicit temporal expressions, it is quite hard to formalize such a proposition in an algorithm, as these events are not bound to a specific syntactic form. Even prepositional phrases such as ``on board'' could be considered an event. Thus, given enough data and computational power, solving such tasks via data-driven models appears to be more feasible.

The problem of sentences with similar meanings being composed in a variety of different syntactic forms was also cited as a reason for the creation of various new annotation frameworks, such as AMR \citep{amr}, which strips syntactic sugar and uses PropBank framesets with pre-defined slots to represent the meaning of a sentence, and UCCA \citep{ucca}, which aims to produce similar annotations for sentences with a similar meaning rather than based on the grammatical structure, by representing them as scenes of processes or states. UCCA's definition of a \emph{scene} is similar to that of an event in TimeML, being composed of either a process that evolves over time, or a state that does not. 

Similar to TCS reasoning, various temporal reasoning domains are also being benchmarked and evaluated using LMs in recent years. For example, \citet{trbenchmarking} evaluate the temporal reasoning capabilities of LMs in closed book QA, open book QA, and reasoning QA formats, finding that such models are often incapable of extrapolating their reasoning capabilities to settings outside the contemporary training period and proposing methods, such as time-sensitive reinforcement learning, to mitigate this issue. As mentioned, we make note of some findings in LM-based temporal reasoning research that can possibly be leveraged in TCS reasoning, but a complete overview is out of the scope of this survey.

\section{Pre-Transformer Temporal Common Sense}
Before providing an overview of modern TCS reasoning models and datasets, we first introduce commonly cited dimensions of TCS reasoning and connect them to previously proposed temporal reasoning tasks. In addition, we showcase some of the main technologies that enabled models to reason over TCS. 

\label{sec:pretransformer}
\subsection{Defining Temporal Common Sense}
The eventual objective of fully automating temporal reasoning is not new. The TimeML authors note the surge in research regarding the automatic recognition of temporal and event expressions in natural language text \citep{time-ml}, posing potential benefits in domains such as question answering. For example, the question ``Did the Enron merger with Dynegy take place?'' requires a model to understand whether an event mentioned in a news article has actually occurred, rather than simply finding any mention of the event.

The TimeBank authors mention that ``from a practical computational perspective, it will become possible to consider training and evaluating algorithms which determine event ordering and timestamping, and to explore their utility in question answering'' \citep{timebank}.

The authors of the TempEval-3 tasks state that the ultimate aim of their research is the ``automatic identification of temporal expressions, events, and temporal relations within a text as specified in TimeML annotation'' \citep{tempeval3}.

However, as introduced in Section \ref{sec:introduction}, such a fully automatic end-to-end pipeline requires models to be able to reason over temporal contexts even when information is only provided implicitly or must be inferred via common sense. Hence, simple data-driven reasoning over explicit contexts is not sufficient to fulfil these visions.

To connect these previous ambitions with current work, we refer to the relatively novel dimensions of TCS as proposed by \citet{mctaco}. These five dimensions are as follows:

\begin{itemize}
    \item \emph{Event typical time}: At what time do we expect certain events to happen?
    \item \emph{Event duration}: How long does an event typically take?
    \item \emph{Event ordering}: What happens before or after a specific event?
    \item \emph{Event frequency}: How frequently does a recurring event typically occur?
    \item \emph{Stationarity}: Does a state hold for a long time or indefinitely?
\end{itemize}

Similar to previous temporal reasoning tasks, these dimensions are also very event-centric. However, the specific phrasing of the dimensions and the resulting reasoning tasks are more open-ended in nature, substituting reasoning over explicit temporal information with a best-effort guess based on common sense. We ask ourselves when an event typically happens or how long it typically takes, but this does not always have to be the case. For example, we may expect most people to shower in the morning, but others may shower in the evening, in the afternoon, or at night. When we talk about TCS, we talk about the average person's expectations for certain temporal properties. It is difficult to explicitly model such concepts, which is why the increased performance of data-driven methods in NLP due to new technologies has been so beneficial to this field.

Notably, the proposed TCS dimensions quite heavily correspond to annotations within TimeML. For example, a temporal link between an event and a timestamp can denote the time at which an event occurs. A temporal link between two events can denote the order of these two events. Certain temporal expressions (e.g., ``on Mondays'') can also signify the frequency of recurring events. Event duration tags for TimeML were also proposed \citep{timeml-duration}. The difference in TCS reasoning is that we do not seek the explicit answer for a specific temporal property in a source text, but rather use pre-existing knowledge to find a likely generalization that applies to an incomplete context.

Systems that possess TCS could thus be expected to understand one or more of these dimensions to reason over downstream tasks. For example, estimating event durations naturally requires systems to know the typical duration of events. Temporal relation extraction mainly requires systems to have knowledge of typical event orderings. Other properties, such as temporal validity \citep{content-expiry-date,tnli,cta}, may require knowledge of a combination of dimensions, such as stationarity and typical event duration.

\subsection{Early Temporal Common Sense Systems}
In the last decade, due to increases in computing power and the evolution of neural models, a focus on semantics helped push the field of TCS reasoning forward. Embedding methods such as \emph{Word2Vec} \citep{word2vec} and \emph{GloVe} \citep{glove} were introduced to generate semantic word embeddings. Although these methods were prone to issues such as a lack of inherent word sense disambiguation, they were used frequently in newer temporal reasoning challenge tasks such as Clinical TempEval \citep{tempeval-2016}. Although the best performing methods still used CRFs and SVMs based primarily on lexical features, neural network structures and static word embeddings were used by many groups, including a proposal for a possible improvement to an RNN-based solution by using the long short-term memory (LSTM) architecture \citep{tempeval2016-rnn}.

In the scope of TCS, \textsc{ROCStories} \citep{rocstories} was one of the first noteworthy datasets to specifically benchmark the understanding of implicit causal and temporal relationships between events in machine learning systems. Although prior work on story comprehension and text understanding existed, for example, in the form of \textsc{MCTest} \citep{mctest}, these datasets did not specifically focus on a temporal or causal aspect nor commonsense reasoning. \textsc{ROCStories}, on the other hand, focused on rich causal and temporal context that was not trivial to resolve, for example, via the sentence order.

Further, \textsc{CaTeRS} \citep{caters} was devised as a new annotation scheme for causal and temporal relations, replacing some TimeML links with causal links based on a previous causal model \citep{causalmodel}. This paper also affirmed the story quality of \textsc{ROCStories} and its temporal properties. \textsc{ROCStories} thus remains an important benchmark dataset in temporal- and causal commonsense reasoning to date.

\section{Modern Temporal Common Sense Benchmarking}
Since 2017, there has been a steady increase in benchmarking datasets for TCS, as well as models aiming to solve the corresponding tasks. In this section, we first briefly describe how the transformer architecture made commonsense reasoning more approachable. We then discuss emerging datasets measuring TCS understanding in LMs and summarize how such datasets may be improved in the future.

\label{sec:sota}
\subsection{Transformer Architecture in Natural Language Processing}
In 2017, the well-known paper ``Attention is all you need'' was published \citep{transformer}, which first introduced the transformer architecture. Over the following years, many LMs based on this architecture would emerge, such as the previously mentioned \textsc{GPT} and \textsc{BERT}, as well as newer models, such as \textsc{GPT-3} \citep{gpt-3} and \textsc{T5} \citep{t5}. The trend with these models is an ever-increasing parameter size and massive amounts of raw text as unsupervised training data, to the point where training their parameters from scratch is often no longer feasible for smaller datasets. Researchers and developers thus often use these models in their pre-trained form, only adding classification layers or extracting the generated word embeddings for downstream tasks and -processing. Another option is to use smaller model sizes, which are less likely to overfit, but may not be able to provide the same reasoning capabilities.

In 2019, a \textsc{BERT}-based model already outperformed existing state-of-the-art systems on temporal relation extraction simply by adding a classification layer on top of the pre-trained model \citep{redcatersbenchmark}. Furthermore, the largest out-of-the-box \textsc{GPT-2} model outperformed state-of-the-art solutions in 7 of 8 evaluated language modelling tasks in a zero-shot setting \citep{gpt-2}. LMs have become so powerful that the largest models do not even have to be fine-tuned to perform specific tasks. For example, \textsc{T5} determines and solves various tasks through a natural language prefix attached to the input, whereas \textsc{GPT-3} can often reason over both the task and corresponding few-shot samples in the input itself. Recently, \textsc{ChatGPT} has shown how prompts, rather than fine-tuning, can be used to solve certain tasks in NLP. However, it is still outperformed by fine-tuned task-specific models on certain tasks, such as sequence tagging \citep{chatgpt-survey}. Naturally, this raises the question of how these models reason over TCS.

\subsection{Temporal Common Sense Benchmark Datasets}
\label{ss:datasets}
Table \ref{tab:datasets} lists TCS benchmark datasets. Except for \textsc{ROCStories}, datasets for TCS benchmarking only began to emerge after the surge in popularity of transformer models. Notably, while the listed datasets are explicitly concerned with TCS reasoning, other temporal reasoning datasets, such as \textsc{TimeQA} \citep{timeQAdataset}, \textsc{MATRES} \citep{matres}, or \textsc{RED} \citep{red-dataset} are sometimes used for benchmarking TCS models as well, although they are not surveyed in this article. We briefly describe the surveyed datasets in the following.

\begin{table}[ht]
    \caption{Temporal common sense benchmark datasets} \label{tab:datasets}
    \begin{tabular}{@{}llllrll@{}}
        \toprule
         Year & Dataset & Task Type & Focus & Size & Context Source & Data Collection\\
         \midrule
         2016 & \textsc{ROCStories} & Classification & No Focus & 50k & Crowdsourcing & Crowdsourcing\\
         2019 & \textsc{Alm2019}\footnotemark[1] & Classification & Duration & 1.7k & \makecell[l]{Wikipedia,\\Blogs, News} & Crowdsourcing\\
         2019 & \textsc{McTaco} & Classification & No Focus & 13k & \textsc{MultiRC} & Crowdsourcing\\
         2020 & \textsc{TORQUE} & Extraction & Ordering & 30.7k & \textsc{TempEval3} & Crowdsourcing\\
         2020 & \textsc{Vash2020}\footnotemark[1] & Classification & \makecell[l]{Duration,\\Ordering} & 1m & \makecell[l]{\textsc{TE3}, \textsc{TB-D},\\\textsc{RED}, \textsc{UDS-T}} & Recasting\\
         2020 & \textsc{WikiHow} & Classification & Ordering & 839k & WikiHow & Crowdsourcing\footnotemark[2]\\
         2021 & \textsc{TIMEDIAL} & Classification & No Focus & 1.1k & \textsc{Dailydialog} & Crowdsourcing\\
         2021 & \textsc{TRACIE} & Classification & Ordering & 5.4k & \textsc{ROCStories} & Crowdsourcing\\
         2023 & \textsc{TNLI} & Classification & Duration & 10.7k & \textsc{Flickr30k} & Crowdsourcing\\
         2023 & \textsc{CoTAK} & Classification & Duration & 300k & WikiHow & Crowdsourcing\\
         \botrule
    \end{tabular}
    \footnotetext[a]{Dataset identifiers are provided by us, as the original authors did not provide explicit names.}
    \footnotetext[b]{Crowdsourcing is only used to train a model to filter articles, not for annotating data instances.}
\end{table}

\textbf{ROCStories}: \textsc{ROCStories} \citep{rocstories} is formulated as a ``story cloze test'', where a model reads the first four sentences of a story and has to choose the correct ending out of two possible options. A rigid crowdsourcing process aims to ensure that stories have sufficient causal and temporal context for a model to choose the correct ending.

\textbf{Alm2019}: This dataset \citep{content-expiry-date} consists of sentences sampled from news, Wikipedia, and blog posts. The authors classify the temporal validity duration of the content. For example, a sentence like ``Joe Biden is the President of the United States'' contains valid information for a longer duration than ``The weather is nice''.

\textbf{McTaco}: \textsc{McTaco} \citep{mctaco} is a multiple-choice question answering dataset that specifically probes all proposed TCS dimensions. Each item contains a short context, such as ``Ratners' chairman, Gerald Ratner, said the deal remains of substantial benefit to Ratners.'', followed by a commonsense question, such as ``How long did the chairman speak?'' A model then has to reason over four possible answer candidates in a binary classification format.

\textbf{TORQUE}: \textsc{TORQUE} \citep{torque} is a reading comprehension dataset focused on temporal ordering. For each text passage, a model must determine which events in the text occur before or after some target event. They focus on a very robust evaluation process to ensure models do not end up scoring high on the dataset through trivial answers.

\textbf{Vash2020}: \citet{Vashishtha2020} recast several event duration and event ordering datasets into a \emph{natural language inference} (NLI) format, in which a given duration of an event or an ordering of a pair of events forms the hypothesis.

\textbf{WikiHow}: \textsc{WikiHow} \citep{wikihow} is a dataset containing steps of WikiHow articles. Among others, they propose a step ordering task. For example, in an article titled ``Clean Silver'', a model would have to determine whether ``dry the silver'' occurs before or after ``handwash the silver''.

\textbf{TIMEDIAL}: Items in \textsc{TIMEDIAL} \citep{timedial}, similar to \textsc{McTaco}, consist of a context and several answer candidates. In contrast to \textsc{McTaco}, \textsc{TIMEDIAL} answer candidates are cloze-style options for a missing temporal quantifier. The contexts themselves are dialogues. An example dialogue is shown below. Models have to evaluate each option in a binary classification format.

\begin{samepage}
    \noindent\rule[1pt]{0.99\textwidth}{1pt}
    
    \nopagebreak
    \noindent A: May we see the wine list, please.
    
    \nopagebreak
    \noindent B: Sure. Our special wine today is a 1989 Chardonnay.
    
    \nopagebreak
    \noindent A: I'd like a bottle, please.
    
    \nopagebreak
    \noindent B: I'll need to see your ID, please.
    
    \nopagebreak
    \noindent A: Here you go.
    
    \nopagebreak
    \noindent
    B: Sorry about the inconvenience, you look so young. I had to make sure you are
    over \rule{0.1\textwidth}{0.3pt}.

    \nopagebreak
    \bigskip
    \noindent\begin{tabular}{@{}lr}
         a) 21 years old & b) 30 years old\\
         c) 4 years old & d) 18 years old
    \end{tabular}\\
    
    \nopagebreak
    \noindent\rule[7pt]{0.99\textwidth}{1pt}
\end{samepage}

\textbf{TRACIE}: \textsc{TRACIE} \citep{tracie}, similar to \textsc{Vash2020}, poses event ordering as a textual entailment task. However, their entailment instances are formulated over intervals rather than discrete points in time, for example: ``event a \emph{starts} before event b \emph{ends}''.

\textbf{TNLI}: \citet{tnli} formulate the \emph{temporal natural language inference} (TNLI) task, in which a model has to determine whether a follow-up sentence supports, invalidates, or is neutral with respect to the temporal validity of actions in the target sentence. For example, the temporal validity of the sentence ``A musician sings into a microphone while playing a guitar.'' is invalidated by the follow-up sentence ``The musician eats at his favourite restaurant.'', as the former action cannot still be ongoing when the latter is observed.

\textbf{CoTAK}: The \textbf{Co}mmonsense \textbf{T}emporal \textbf{A}ction \textbf{K}nowledge dataset \citep{cta}, similar to \textsc{WikiHow}, contains steps of WikiHow articles. Based on the heading of a given step (and possibly the title of the article itself), the goal is to predict the \emph{action perform duration}, which is effectively the typical event duration, as well as the \emph{action effect duration}, which the authors equate with the temporal validity duration of the action.

\subsection{Categorization of Benchmarked Datasets}
Several traits listed in Table \ref{tab:datasets} can be more closely analysed. We observe two distinct categories of task types, \emph{classification} and \emph{extraction}, in the surveyed datasets. We denote a task as a \emph{classification} task if a model is expected to reason over the likelihood of a set of provided answer candidates. This is the case in most proposed datasets. The actual specific task varies, as seen in Section \ref{ss:datasets}. Of the proposed datasets, only the questions in \textsc{TORQUE} are more open-ended in the form of an extractive question answering task. However, this task can also be resolved by a binary classification of each token in the text passage. 

\begin{figure}[htbp]
    \centering
    \includegraphics[width=119mm]{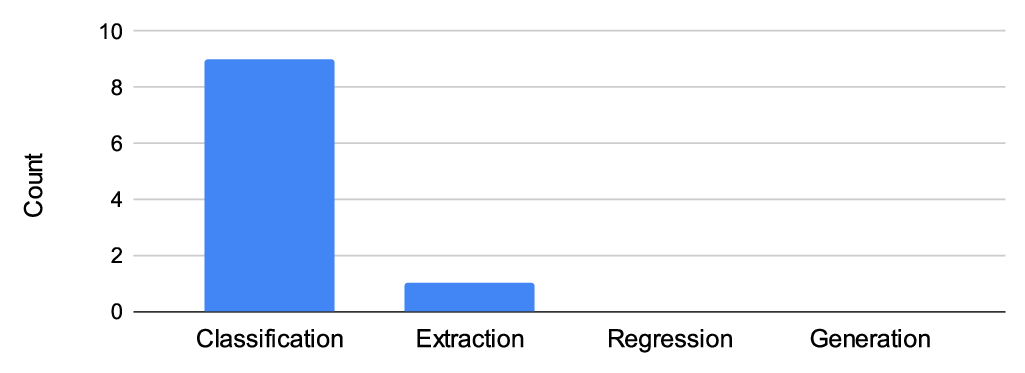}
    \caption{Distribution of task types in surveyed datasets}
    \label{fig:dscategories}
\end{figure}

Notably, further categories might include \emph{regression} or \emph{generation} tasks, but these are so far absent from the surveyed datasets. Only a few articles, such as \citet{timeawarept}, aim to model certain dimensions of existing datasets, such as event duration estimation, as regression tasks, but this can lead to mismatches between the intended task and the evaluated results. For example, by mapping \textsc{McTaco} answer candidates from text to a normalized duration for regression comparison, an evaluation of the model's ability to reason over different representation of the same interval (e.g., ``90 minutes'' versus ``1.5 hours'') is lost. For this reason, datasets that actively encourage different label representations could be much more effective for benchmarking such approaches. Likewise, datasets for generative approaches, in which a model would autonomously provide a reasoning for giving a certain answer or provide its best effort guess without answer candidates, may more closely model the current trend of foundation models and prompt-engineering. In Figure \ref{fig:dscategories}, we visualize these trends in a graphic, highlighting the need for datasets suitable for different output formats.

\begin{figure}[htbp]
    \centering
    \includegraphics[width=119mm]{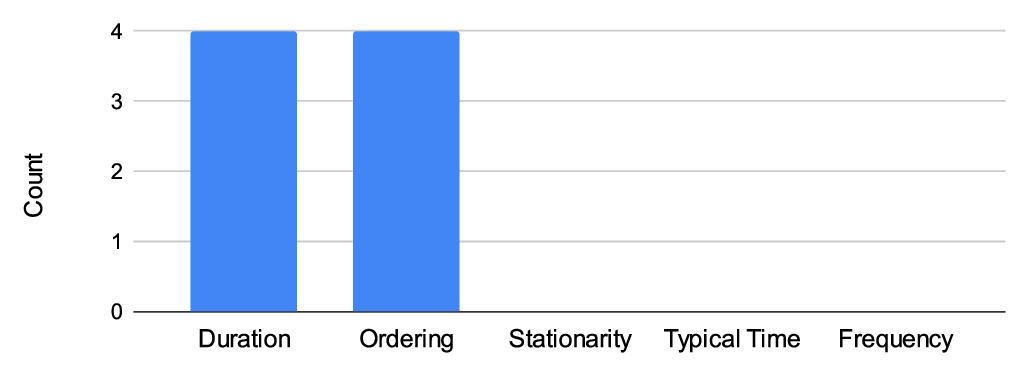}
    \caption{Distribution of TCS focus dimension categories in surveyed datasets (only datasets that have a focus)}
    \label{fig:focusdimensions}
\end{figure}

As for the TCS focus dimension being benchmarked, it is usually event ordering or event durations. In other cases, there may be no specific categorization of dimensions, and TCS is presumed to be measured in its entirety. Given the discussed history of temporal reasoning and how it ties into newer TCS research, the focus on event ordering and -durations is unsurprising, as these dimensions most closely model previous temporal reasoning tasks. The three datasets \textsc{McTaco}, \textsc{TIMEDIAL}, and \textsc{ROCStories} mainly focus on contextual understanding of the temporal properties of a given background story without considering the specific commonsense dimensions too closely. However, the authors of \textsc{McTaco}, as their paper defines the dimensions discussed earlier, categorize each question into one of the five dimensions. Although it could be argued that temporal validity tasks, such as those proposed by \citet{content-expiry-date} or \citet{tnli}, also require reasoning over stationarity, in existing literature, reasoning over stationarity is mostly circumvented by filtering stationary data from the dataset. However, this filtering in itself can be a difficult process, especially due to the lack of state-of-the-art models that reason over the stationarity property, which further highlights the need for new datasets that can be used to train and benchmark models on the remaining TCS dimensions. The focus dimension distribution is shown in Figure \ref{fig:focusdimensions}, for all datasets that have at least one focus dimension.

Also of note is that the proposed datasets rely heavily on crowdsourcing. Except for \textsc{Vash2020}, all datasets use crowdsourcing during dataset construction. Authors of the \textsc{WikiHow} dataset use crowdsourcing to train a \textsc{BERT}-model to predict whether the steps in articles are ordered for downstream processing, but not for determining the commonsense properties of the texts themselves. All remaining authors create their dataset at least partially via crowdsourced annotations. Several authors also source contexts from existing datasets, which in turn may also have been created by crowdsourcing. It is relatively logical to rely on crowdsourcing to create datasets for commonsense reasoning purposes, but it can be prone to errors or fraudulent activity by workers, which potentially requires manual intervention to ensure that the dataset quality remains high \citep{tnli}. Platforms such as \textsc{CrowdAQ} \citep{crowdaq} can potentially help to properly train and vet crowdworkers for specific tasks. However, it has been shown that items in common sense datasets often do not stand up to expert vetting regardless \citep{benchmark-survey}, which can be an issue when those datasets are used to benchmark model performance.

The size of the proposed datasets also varies significantly, and the average dataset size is relatively small. However, the authors of \textsc{TORQUE} show that this may not always be an issue, as the performance of their baseline approach converges before much of the available training data is ingested into the model \citep{torque}. As most datasets only aim to benchmark performance rather than teaching new reasoning capabilities to a model, the model only has to be post-trained enough to understand the given task format. 

There are two significant outliers with respect to size. \citet{Vashishtha2020} create a considerable number of NLI pairs by recasting existing temporal relation extraction datasets into an NLI format. For example, for the phrase ``We waited until 2:25 PM and then left'', we can formulate a hypothesis such as ``The waiting started before the leaving started'', for which the answer is known from existing annotations. However, it should be considered that, in this case, not all samples are guaranteed to measure TCS understanding, as the answer may already be provided explicitly by the statements in question. Similarly, \citet{wikihow} infer the step ordering from the WikiHow articles directly, after first using the previously mentioned \textsc{BERT}-model to determine whether the article contains ordered steps. \textsc{CoTAK} is the largest dataset with crowdsourced target labels, at over 300,000 samples.

\subsection{Lessons from Existing Benchmarking Datasets}
\label{subsec:improveds}
For the remainder of this section, the goal is to summarize the dataset authors' reflections and draw parallels with work in related fields. We list important findings and proposed methods to improve the robustness of future work.

\subsubsection{Evaluation Metrics} 
When reporting on classification tasks, commonly reported quantitative metrics, such as accuracy and F1 score, are generally the most generous interpretations of model performance. A model can achieve high accuracy or F1 score simply by exploiting patterns in the data. The most straightforward example is, of course, a model that simply predicts the majority class. In binary classification settings (such as multiple choice question answering, where every question-answer pair resolves to either true or false), such a system is guaranteed to obtain an accuracy score of at least 0.5. Although the F1 score is somewhat more robust (assuming that the problem has a reasonable class distribution), a model can still achieve a high F1 score by finding simple patterns in the dataset without actually understanding the problem statement.

Thus, using a context-level \emph{exact match} (EM) metric may be preferable, where applicable. The rationale is that a system that can genuinely reason over a specific property (such as TCS) should be evaluated on the number of contexts it can reason over flawlessly (e.g., the number of questions for which a system can correctly classify all possible answers as true or false). On the other hand, a metric like accuracy measures performance on a case-by-case basis and disregards consistency within the model.

For example, suppose that a system can identify ``1.5 months'' as a correct answer to a given question but not ``6 weeks''. In that case, likely, the system is not using TCS to arrive at this conclusion, but is instead taking a shortcut in the reasoning, such as pattern matching one of the proposed answer candidates. The datasets \textsc{McTaco} and \textsc{TRACIE} provide EM scoring at the context level in their baseline performance reporting. \textsc{TIMEDIAL} is evaluated on \emph{2-best accuracy}, in which both correct answer candidates must be ranked as more likely than both incorrect answer candidates, which provides somewhat of a middle ground between accuracy and context-level EM.

\subsubsection{Contrast Sets} To even further decrease the likelihood of overvaluing incidental correct responses, \citet{contrastsets} propose creating contrast sets. These datasets contain data points that are as similar as possible to data points within the dataset, while leading to a different classification result. They show that while humans succeed on such contrast sets, models often do not, specifically mentioning \textsc{McTaco} as an example, which drops from 0.38 EM to 0.14 EM on such a contrast set. In the example below \citep{contrastsets}, the contrast instance differs only in a few words from the original context, but drastically changes the expected likelihood of the candidate answer.

\begin{samepage}
    \noindent\rule[1pt]{0.99\textwidth}{1pt}

    \nopagebreak
    \noindent \textbf{Context}: She renews in Ranchipur an acquaintance with a former lover, Tom Ransome, \emph{now a dissolute alcoholic}.
    
    \nopagebreak
    \noindent\textbf{Contrast}: She renews in Ranchipur an acquaintance with a former lover, Tom Ransome, \emph{who keeps very healthy habits}.
    
    \nopagebreak
    \noindent\textbf{Question}: How frequently does Tom drink?
    
    \nopagebreak
    \noindent\textbf{Candidate Answer}: Every other night.
    
    \nopagebreak
    \noindent\rule[3pt]{0.99\textwidth}{1pt}
\end{samepage}

The authors of \textsc{TORQUE} try to implement this measure by specifically negating their temporal ordering questions to maximize the difference in the desired output. For example, if a question in the dataset is ``What happened after he ate his breakfast?'', the contrast questions ``What happened when he was eating his breakfast?'' and ``What happened before he was eating his breakfast?'' should also be posed. In total, the answer to these questions should cover all possible events in the context, and each event should resolve only to the correct question. They then report EM consistency, the percentage of contrast question sets for which a model's predictions match exactly.

To highlight the impact of reporting more robust metrics, we list some evaluation metrics reported from the baseline models for \textsc{McTaco} and \textsc{TORQUE} in Table \ref{tab:metrics}. Here, the previously discussed EM consistency is denoted as C for TORQUE. These results strongly highlight that model performance decreases much more than human performance on such metrics.

\begin{table}[ht]
    \caption{A list of reported evaluation metrics in \textsc{McTaco} and \textsc{TORQUE}, sorted by performance difference between humans and the best-performing model presented in the paper ($\Delta$)}
    \label{tab:metrics}
    \begin{tabular}{@{}llrrr@{}}
        \toprule
         Dataset & Metric & Model & Human & $\Delta$\\
         \midrule
         \textsc{McTaco} & F1 & .699 & .871 & .172\\
         \textsc{TORQUE} & F1 & .752 & .953 & .201\\
         \textsc{McTaco} & EM & .427 & .758 & .331\\
         \textsc{TORQUE} & EM & .511 & .845 & .334\\
         \textsc{TORQUE} & C & .345 & .825 & .480\\
         \botrule
    \end{tabular}
\end{table}

\subsubsection{Measuring Model Understanding}
Although humans tend to succeed more than TCS models when evaluation metrics are stricter, dataset authors should take care to report metrics that aim to measure the model's understanding of the problem as accurately as possible. In extractive question answering, token-level F1 and EM score are two metrics that are typically used for evaluation. However, unlike classification problems, where answers are unambiguously correct or incorrect, there is often some ambiguity when it comes to extracted text spans. 

The authors of \textsc{TORQUE} provide token-level F1 and EM score, rather than context-level. In their dataset, this appears not to pose a problem, as their questions are effectively a natural-language recasting of known temporal relations between events. However, this does not apply to every problem. \citet{answerequivalence} note that both token-level F1 and EM fail to recognize cases in which a model may remove incorrect information or add further relevant information to its response due to the symmetry of the metrics. They propose an asymmetric \emph{answer equivalence} metric, as well as a \textsc{BERT}-based estimator for said metric. Although they concede that their approach does not address a potential temporal dimension of answer candidates (e.g., ``4 months ago'' and ``February 2022'' is only equivalent in June 2022), this asymmetric type of answer scoring may provide a better solution for extractive tasks and may pave the way for eventual generative approaches, where a generated answer candidate could be reasonably compared against the reference answer.

In summary, dataset authors should take care in their baseline reporting to identify and propose appropriate evaluation metrics. Where possible, these metrics should discourage models from obtaining high scores simply by finding a reasoning shortcut for a subset of the data. They should be as strict as reasonably possible while still allowing the model to be expressive in its responses, should the task format allow for it. Although classification tasks, which form the basis of many currently existing datasets, inherently do not pose a risk of restricting the expressiveness of the model, they also do not always align closely with downstream tasks, since answer candidates are not always available. For more autonomous models that do not rely on answer candidates, different metrics will thus need to be used.

\subsubsection{Linguistic Traps} 
The authors of \textsc{TimeQA} showcase ``shallow pattern matching'' performed by transformer models through their split of \emph{easy} and \emph{hard} questions. For example, if an athlete was on a team between 1973 and 1975, an easy question might be, ``What team did [player] play for between 1973 and 1975?'' A hard question might instead be ``What team did [player] play for in June 1974?'' or ``What team did [player] play for between April 1974 and December 1974?'', since the exact temporal spans are not reused. While human annotation only incurs a performance decline of 2 percent in the EM metric from such questions, it is roughly 13.7 percent for the best-performing transformer system.

This dependence on simple pattern matching can also be seen in the \textsc{TIMEDIAL} dataset. Here, crowdworkers were explicitly instructed during dataset creation to try to reuse explicit temporal quantifiers from the question in incorrect answer candidates wherever possible. They show that a simple \textsc{BERT} model picks such incorrect options 52 percent of the time over the correct answers. For example, when the context mentions a meeting starting at ``three o'clock'' for which the speaker does not want to be late, models were more likely to estimate ``half past three'' as a possibility for the current time than ``quarter to two''.

Although more powerful transformer models, such as \textsc{T5}, are more robust to pattern matching, it remained the most common error type. It can also be linked to previously reported issues such as mispriming. For example, a \textsc{BERT} model may fill the mask in ``Samsung. The iPhone is produced by [MASK]'' with ``Samsung'' due to the previous mention of the in-domain phrase. Similarly, other pitfalls, such as ignoring negation and word order, have also been reported in transformer-based LMs \citep{pitfalls}.

\subsubsection{Debiasing} 
While not TCS research, the WinoGrande \citep{winogrande} paper highlights the importance of debiasing in NLP. They create a crowdsourced dataset for commonsense pronoun disambiguation and evaluate fine-tuned models on the well-known Winograd Schema Challenge dataset \citep{winograd}. Although their own WinoGrande dataset was crowdsourced, they achieved better performance on the original expert-crafted dataset, citing their debiasing strategy as a critical reason for this success.

Debiasing generally comes in the form of some adversarial learning, which ensures that solving the instances in the dataset is not trivial for the model. In the case of WinoGrande, this is done by using linear classifiers based on \textsc{RoBERTa} \citep{roberta} embeddings to remove the easiest instances to classify from the dataset repeatedly. 

The authors of the \textsc{CIDER} common sense dataset, which also has some temporal dimensions, apply adversarial filtering to remove stylistic patterns from their confounding-option candidates.

\section{Improving Temporal Commonsense Reasoning}
\label{sec:approaches}
In the previous section, we show how TCS datasets can be constructed, as well as some techniques that can make their evaluation more robust to typical strategies exploited by modern LMs. In this section, we discuss proposed methods to improve the current state of the art in TCS reasoning.

\subsection{Baseline Models and Human Performance}
\label {subsec:baselines}
First, we examine the results that the dataset authors provide as baselines. Table \ref{tab:baselines} shows the performance of baseline models. Note that the provided results in Table \ref{tab:baselines} are not necessarily the best result reported in the paper, as we will examine the impact of augmentations in a later section, but that the results do showcase the performance of the best model structure in its base form. \textsc{ROCStories} was published before the transformer architecture was popularized. As such, the authors report performance by another neural network architecture called DSSM (deep semantic similarity model). Despite being published in 2019, the authors of \textsc{Alm2019} do not report the performance of any transformer-based model, instead using a support vector classifier (SVC). \textsc{TNLI}'s \textsc{SelfExplain} baseline uses the model by \citet{selfexplain}, whose embeddings are also based on RoBERTa. However, \textsc{SelfExplain} adds some additional layers which greatly improve performance on \textsc{TNLI}.

\begin{table}[ht]
    \caption{Performance of baseline models reported by dataset authors}
    \label{tab:baselines}
    \begin{tabular}{@{}lllrr@{}}
        \toprule
        Dataset & Base Model & Metric & Performance & Human\\
        \midrule
        \textsc{ROCStories} & DSSM & Acc & .585 & None\rlap{\footnotemark[2]}\\
        \textsc{Alm2019} & SVC & F1 & .702 & None\\
        \textsc{McTaco} & \textsc{BERT} & F1 & .699 & .871\\
        \textsc{TORQUE} & \textsc{RoBERTa-large} & F1 & .752 & .953\\
        \textsc{Vash2020} & \textsc{RoBERTa-large} & Acc & .809\rlap{\footnotemark[3]} & None\\
        \textsc{WikiHow} & \textsc{RoBERTa} & Acc & .835 & .975\\
	    \textsc{TRACIE} & \textsc{RoBERTa-large} & Acc & .784 & .825\rlap{\footnotemark[4]}\\
        \textsc{TIMEDIAL} & \textsc{T5-large} & 2-best Acc & .748 & .978\\
        \textsc{TNLI} & \textsc{SelfExplain} & Acc & .873 & None\\
        \textsc{CoTAK} & \textsc{BERT-base-uncased} & Acc & .775\rlap{\footnotemark[5]} & None\\
        \botrule
    \end{tabular}
    \footnotetext[b]{Authors report a human performance of 1.0 due to double-verified crowdsourced instances, but no further human evaluation was performed.}
    \footnotetext[c]{Macro-average performance across all evaluated datasets by the overall best-performing model.}
    \footnotetext[d]{Reported on a \emph{more difficult} (compared to model performance) ``no-story'' set.}
    \footnotetext[e]{Macro-average of coarse-grained duration estimates based on only the action description.}
\end{table}

In the \textsc{TRACIE} dataset, human performance is only reported for a ``no-story'' set, in which only the hypothesis of a story is known, but not its context. On the other hand, the reported model performance is based on the dataset that contains complete information. Hence, humans had to solve a more difficult task, as they had much less available information. Nevertheless, they still outperformed the best-performing baseline model.

\subsection{Proposed Augmentations}
\begin{figure}[htbp]
    \centering
    \includegraphics[width=119mm]{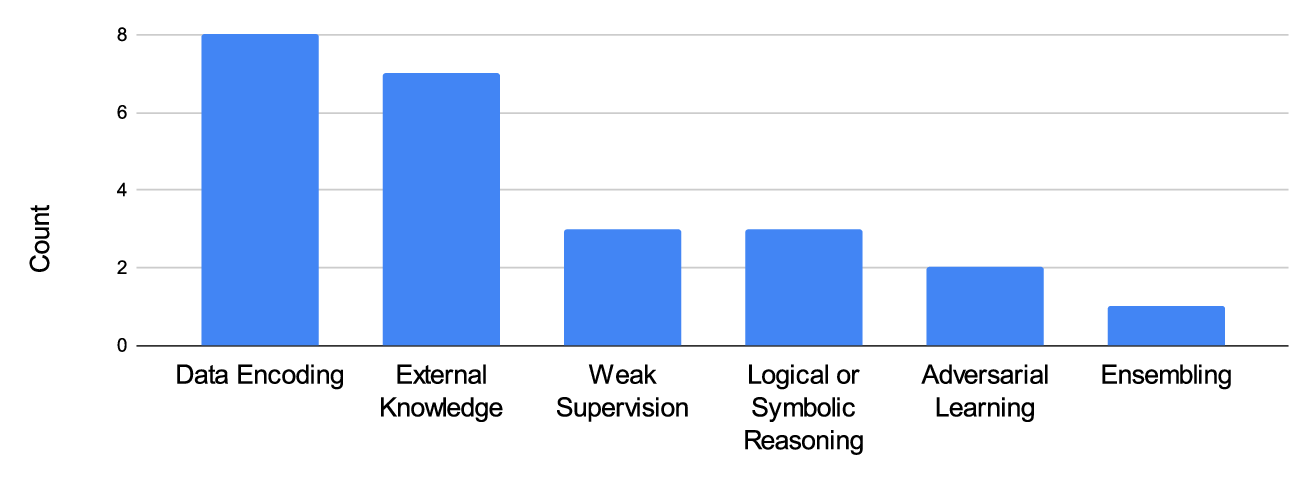}
    \caption{Distribution of observed augmentation categories in the core literature}
    \label{fig:augmenttypes}
\end{figure}

Notably, none of the baseline models proposed in Section \ref{subsec:baselines}  appear to understand temporal attributes well enough to match human performance. Consequently, their out-of-the-box reasoning over TCS dimensions can still be improved. As noted in Section \ref{sec:collection}, augmenting existing transformer-based LMs is a common approach in TCS reasoning. Figure \ref{fig:augmenttypes} shows how often we observe specific augmentation types in the surveyed literature. Data encoding and external knowledge sources appear to be some of the most frequently observed augmentations, but other methods, such as logical or symbolic reasoning and adversarial learning, have also led to performance improvements that should not be overlooked. In the remainder of this section, we explore the proposed techniques to improve the TCS understanding of LMs in more detail.

\subsubsection{External Knowledge}
It is likely that a significant reason for the lack of commonsense understanding in transformer models is reporting bias. Due to the nature of language, using the frequency of event occurrences in text as a baseline for commonsense knowledge is generally not ideal \citep{reportingbias-general}. Also known as the ``black sheep problem'', we intuitively understand that one is much more likely to mention a ``black sheep'' than to specify the colour of a regular sheep, which may confuse statistical models. These artefacts can be seen in the event likelihood estimate of transformer models. For example, \textsc{BERT}, which was trained on Wikipedia, may overestimate the likelihood of death. Similarly, \textsc{RoBERTa}, which was trained on the web, overestimates the probability of newsworthy events such as being murdered \citep{reportingbias}.

Several recent papers have attempted to mitigate this bias by using KGs with LMs. Specifically, the two previously mentioned KGs \textsc{ConceptNet} and \textsc{ATOMIC}$^{20}_{20}$ have frequently been proposed for such methods due to their specific temporal relations (e.g., ``X causes Y'', or ``After doing X, person Y will want to…''). KGs can be used in TCS models to provide ``knowledge embeddings'' of phrases \citep{tnli}, or to directly post-train the LM on KG triples converted to natural language \citep{commonsense-storygen}. 

The \textsc{CoCoLM} model \citep{cocolm} uses the \textsc{ASER} KG for pre-training. Unlike the KGs mentioned above, \textsc{ASER} triples are automatically constructed from raw text, which means that the KG contains more instances, but may contain noise. \textsc{CoCoLM} shows significant gains on \textsc{ROCStories} using a base \textsc{BERT} model and random walk over \textsc{ASER} to generate multi-hop reasoning phrases as training instances. 

Although knowledge graphs currently appear to be the main source of external commonsense information in TCS, other sources, such as script knowledge systems \citep{frumpsk}, could also be bootstrapped similarly, including possibly synthetically generating such scripts via LMs.

Another source of external knowledge can also be out-of-domain tasks, including multitask training on datasets from related domains (e.g., temporal reasoning), or auxiliary tasks that provide additional information that is relevant to the main task. For example, \textsc{TacoLM} \citep{taco-lm} is a \textsc{BERT}-based model that predicts temporal upper bounds for events, as well as relative hierarchies between events, to improve its TCS reasoning. \textsc{CoCoLM} also introduces auxiliary tasks, such as discourse relation prediction, to improve performance on the \textsc{ROCStories} dataset.

\subsubsection{Weak Supervision}
Weak supervision in TCS reasoning is often based on event co-occurrences with temporal expressions, which can be used to train an LM. \citet{content-expiry-date} propose using the co-occurrence of temporal expression with subjects, verbs, objects, and their combinations as a feature in an SVM-based classifier. A similar approach was proposed in the form of \textsc{TemProb}, a statistical knowledge base showing common relations between events extracted from 20 years worth of NYT articles \citep{temprob}.

\textsc{TacoLM} also uses syntactic rules to extract large quantities of event durations from text, which can then be used as labels for the previously mentioned auxiliary tasks. The resulting model can predict the duration and frequency of events much better than a standard \textsc{BERT} model and has considerably more TCS knowledge. 

Similarly, \citet{timeawarept} extract events and their corresponding duration expressions using rule-based patterns, and use them as weak supervision labels to train a regression-based model for event duration estimation.

The previously discussed \textsc{CoCoLM} model also uses weak supervision, as instances in \textsc{ASER} are automatically extracted in such a manner. Weak supervision can be powerful, as there is plenty of raw text from which large datasets can be created. However, only high-precision patterns should be used to automatically extract information from raw text, as any noise can significantly hinder the training objective. For example, the automatically generated \textsc{ASER} KG only outperforms \textsc{ATOMIC}$^{20}_{20}$ as a knowledge source in \textsc{CoCoLM} when multi-hop reasoning is used to generate training phrases, but not when the model is trained directly on its triples \citep{cocolm}.

\subsubsection{Symbolic or Logical Reasoning}
Another approach is the introduction of symbolic or logical reasoning into commonsense models. The \textsc{SymTime} model \citep{tracie} is an example of symbolic reasoning. An encoder-decoder model estimates the duration of an event and the distance between two events into a set of classes. The softmax distribution of the duration and distance estimates is then used to symbolically reason over the feature vectors to determine whether the estimated duration of event A is longer than the estimated distance between event A and event B. This information is then used to solve the event ordering task. Here, the relationship between duration and distance is explicitly modelled, rather than relying on an LM to learn it implicitly.

Another example is the \textsc{SLEER} model \citep{sleer}, which also explicitly models the relationship between temporal dimensions in the form of logical propositions. An example of such a proposition is as follows:

\begin{quote}
    DUR(\emph{e1, year}) $\Rightarrow$ FREQ(\emph{e1,decade}) $\vee$ FREQ(\emph{e1,century})
\end{quote}

This proposition states that an event with a duration span of year(s) cannot occur more than yearly, and thus must have a frequency of either decades or centuries. The \textsc{SLEER} model uses probabilistic soft logic to express the truths of such propositions on a continuous scale. The distance between expected true statements and the prediction of the model can then be used as a parameter in the loss function to train the LM.

The recently proposed \textsc{LECTER} model \citep{ssl-induction} similarly uses a combination of temporal expression defuzzifying together with probabilistic logic programming to greatly improve the performance on \textsc{TIMEDIAL}. After normalizing and embedding temporal expressions, a logic induction layer generates the probability distribution of relationships between said expressions, and DeepProbLog is used to apply logical entailment to the loss function of the model. With the symbolic temporal logic induction module, \textsc{LECTER} may also be more explainable than common LM-based methods.

In general, explicitly leveraging such logical relationships seems to improve the results of the corresponding LMs. This may indicate that explicitly coding our understanding of relationships between temporal dimensions into reasoning models may outperform implicitly encoding them in LMs using auxiliary tasks.

\subsubsection{Information Encoding}
Going beyond the standard token-level text encoding that transformers usually leverage may also be helpful in some instances. For example, \citet{eventoptimaltransport} propose an approach to modelling text on an event level rather than a token level. Additionally, they propose \emph{event optimal transport} (EOT) as a loss function to better align texts where a regular token-level similarity may lead to poor results. For example, ``Investors bought stocks'' may be considered a better approximation of ``Investors sold stocks'' on a token level than ``British investors sold stocks'', but event-based encoding and event optimal transport help identify similar events and event orders even when they are not aligned. They show that this approach performs well on event ordering and event infilling tasks.

In temporal reasoning, researchers also commonly consider how time can be embedded in an LM. Methods such as prepending a time-specific token to the input \citep{tempwikibio}, altering the transformer architecture directly to temporalize the attention mechanism \citep{tempattention}, or masking and predicting temporal expressions or the document timestamp \citep{tempobert,timebert} have been proposed.

In general, the ideal encoding of a model also somewhat depends on the downstream task and the available information. For example, online content such as news articles or blog posts may contain more readily available document creation date information. On the contrary, such a model may fail on the narratives proposed in \textsc{ROCStories}, which do not contain explicit timestamps. 

\subsubsection{Adversarial Learning}
The adversarial augmentation proposed in Section \ref{subsec:improveds} can also occur at the model level, as shown by the \textsc{ALICE} \citep{alice} and \textsc{ALICE++} \citep{alicepp} models. In these models, the inputs are minimally perturbed during training to maximize the predicted change in the output. In \textsc{ALICE++}, this perturbation additionally occurs on layers besides the input, up to some top layer of the model. Through these engineered samples, the robustness of the model to small changes increases. In practice, this learning method appears to be very effective. For example, \textsc{ALICE++} outperforms the previously mentioned \textsc{SymTime} model, which is based on \textsc{T5}, on the \textsc{MATRES} dataset, despite using \textsc{RoBERTa} for its training, which is a much smaller LM. It also outperforms models such as \textsc{TacoLM} in datasets like \textsc{McTaco}. Overall, similar to adversarial samples in the dataset, this type of learning can enhance performance on the model level as well.

\subsubsection{Ensembling}
Finally, using a combination of multiple classifiers can also enhance model performance. In the TCS domain, the performance on \textsc{McTaco} was improved by constructing an ensemble of multiple BERT models, each fine-tuned on different datasets, using a majority vote to determine the final class \citep{ensemble-bert-tcs}.

This ensemble approach is intriguing, not just because of the growing number of benchmarking datasets for TCS, but also for its potential to evaluate combinations of the augmentations proposed in the rest of this section. Rather than performing an ablation study of the different augmentation types on a single model, evaluating ideal weights for an ensemble classifier of different models, each enhanced with a specific augmentation type, is another option to provide further insight into the value of each model type and possibly improve performance on TCS tasks.

\section{Discussion}
\label{sec:discussion}
In this section, our goal is to highlight the results of our survey and propose possible future research opportunities.

\subsection{Defining and Benchmarking Temporal Common Sense}
In this survey, we have highlighted several similarities between early temporal algebras, temporal reasoning tasks such as temporal relation extraction and event ordering, and proposed TCS reasoning dimensions. We pose that the main difference between TCS reasoning and other temporal knowledge which a model may have (such as temporal factual knowledge or reasoning capabilities over an explicit temporal context) is the inherently probabilistic nature of common sense. While we can make assumptions about the likely order, duration, or time of occurrence of individual actions, common sense does not make any guarantees. By design, the context in a TCS task does not give us a concrete answer, but we should use our prior understanding of the world to derive a likely one. We can consider TCS reasoning to be a probabilistic recasting of existing temporal reasoning tasks.

Based on our survey, we propose the following future work for training and benchmarking TCS.

\begin{itemize}
    \item Datasets focusing specifically on \emph{typical event times}, \emph{stationarity}, and \emph{event frequency} would help improve the general TCS understanding of models. While training data for typical event order and duration can often be derived from text or existing temporal reasoning datasets, this has not been attempted as frequently for the three remaining dimensions.
    \item Regardless of the focus dimension, where possible, explicitly stating which type of TCS is expected (e.g., event ordering or event duration) can help other researchers better identify tasks or benchmarks that a specific model should be able to solve. While the general question of how much TCS understanding transformer models possess is interesting, many downstream tasks may not require a model to be able to reason over all proposed dimensions. For example, for a sentence ordering task, the typical order of events is much more important than the typical frequency of an event.
    \item Most TCS reasoning datasets pose a closed-ended QA or NLI task, which is almost always solved via binary classification (either the candidate answer fits or it does not fit). However, for downstream tasks, different task formats, such as ordinal classification, extractive question answering, regression, or text generation, could be beneficial in providing more detailed training and evaluation data. Additionally, a model can better learn different temporal properties, like how long an action is actually expected to take, rather than simply understanding if certain predetermined answer candidates apply.
    \item When creating new datasets or evaluating a model on existing ones, care should be taken that relatively simple metrics (such as accuracy) do not skew the perceived capabilities of the model. Contrast sets and exact match metrics can help ensure that a machine learning model can genuinely reason over a set of items, rather than relying on a shortcut that the model may have found to distinguish between the target classes. On the other hand, extractive and generative models should have the freedom to deviate from reference answers, as long as the provided answer still solves the problem. 
\end{itemize}

\subsection{Improving Temporal Commonsense Reasoning}
\begin{figure}[htbp]
    \centering
    \includegraphics[width=119mm]{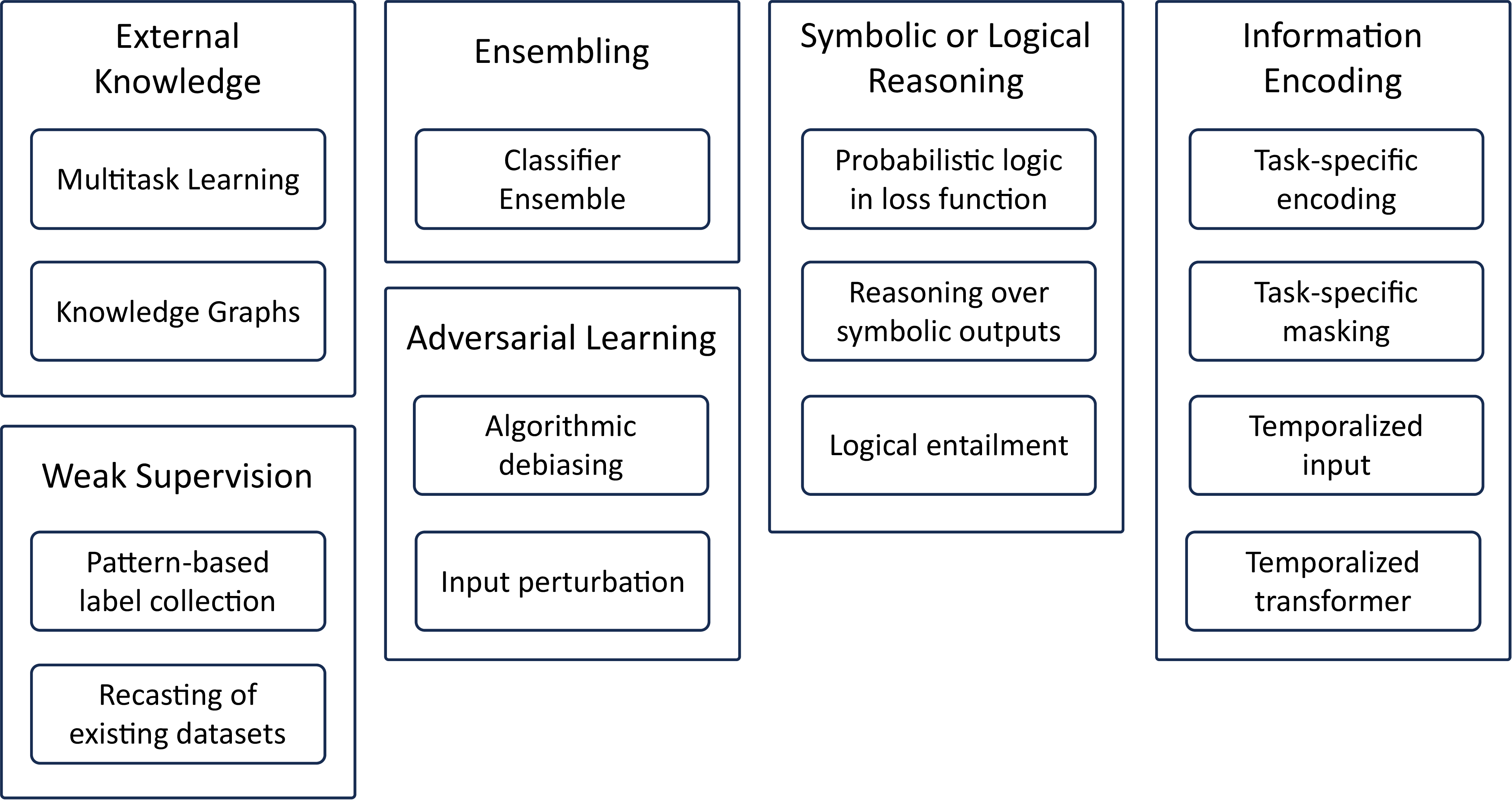}
    \caption{A summary of the augmentation categories and the implementations that were observed}
    \label{fig:augcategories}
\end{figure}

We have discussed proposed augmentations for the transformer architecture to improve TCS reasoning. Table \ref{tab:augmentations} shows an example for each of the proposed augmentation categories, as well as the resulting improvement in performance over their respective base models. We also summarize the advantages and disadvantages of the proposed augmentation types in Table \ref{tab:augsummary}, and show an overview of augmentation categories and specific observed implementations of said categories in Figure \ref{fig:augcategories}. 

\begin{table}[htbp]
    \small
    \caption{Sample augmentations to transformer models and their performance impact. (EXT - External Knowledge; ENC - Data Encoding; LSR - Logical or Symbolic Reasoning; ADV - Adversarial Learning; WS - Weak Supervision; ENS - Ensemble)}    
    \label{tab:augmentations}
    \begin{tabular}{@{}lllcrrl@{}}
        \toprule
         Dataset & Base Model & Cat. & Metric & Base & Aug. & Ref.\\
         \midrule         
         \textsc{TNLI} & \textsc{SelfExplain} & EXT & Acc & .873 & .878 &\citet{tnli}\\         
         \textsc{McTaco} & \textsc{BERT} & WS & EM & .421 & .427 &\citet{taco-lm}\\
         \textsc{TRACIE} & \textsc{RoBERTa}-large & LSR & F1 & .784 & .806 &\citet{tracie}\\
         \textsc{McTaco} & BART-large & ENC & F1 & N/A & .623 &\citet{eventoptimaltransport}\\
         \textsc{McTaco} & \textsc{RoBERTa}-large & ADV & EM & .511 & .599 &\citet{alicepp} \\
         \textsc{McTaco} & \textsc{BERT} & ENS & EM & .396 & .465 &\citet{ensemble-bert-tcs}\\
        \botrule
    \end{tabular}
\end{table}

\begin{table}[ht]
    \caption{Summary of advantages and disadvantages of different augmentations}
    \begin{tabularx}{\textwidth}{lX}
        \toprule
        Augmentation Type & Pros \& Cons\\
        \midrule
        External Knowledge & \begin{minipage}[t]{\linewidth}\textbf{Advantages}\\ - Exposes the model to a large quantity of (typically human-verified) knowledge that may be infrequently observed in text.\\ \textbf{Disadvantages}\\ - Requires a (usually large, often manually constructed) external knowledge source.\end{minipage} \\
        \addlinespace
        \hline
        \addlinespace
        Weak Supervision & \begin{minipage}[t]{\linewidth}\textbf{Advantages}\\ - Can efficiently generate a large amount of additional training data for the primary or auxiliary tasks.\\ \textbf{Disadvantages}\\ - Requires a high-precision approach for extracting labels.\\ - Different methods may be more or less sensitive to noisy labels.\end{minipage} \\
        \addlinespace
        \hline
        \addlinespace
        Symbolic and Logical Reasoning & \begin{minipage}[t]{\linewidth}\textbf{Advantages}\\ - Resulting labels may be more well-founded and explainable than other methods.\\ - The interplay between temporal dimensions can be exploited.\\ \textbf{Disadvantages}\\ - Sound rules must be defined and translated into a format that is appropriate for the architecture.\end{minipage} \\
        \addlinespace
        \hline
        \addlinespace
        Information Encoding & \begin{minipage}[t]{\linewidth}\textbf{Advantages}\\ - Often comparatively simple to implement.\\ - Many different approaches that can be task-specific or task-agnostic.\\ \textbf{Disadvantages}\\ - The resulting performance gains may not be as explainable as other methods.\end{minipage} \\
        \addlinespace
        \hline        
        \addlinespace
        Adversarial Learning & \begin{minipage}[t]{\linewidth}\textbf{Advantages}\\ - Can partially counteract some shallow reasoning behaviours in transformers.\\ - May make the model less sensitive to outliers.\\ \textbf{Disadvantages}\\ - Requires synthetic generation of adversarial samples.\end{minipage} \\
        \addlinespace
        \hline
        \addlinespace
        Ensembling & \begin{minipage}[t]{\linewidth}\textbf{Advantages}\\ - Relatively simple to implement.\\ \textbf{Disadvantages}\\ - More computationally heavy to train and generate target labels for multiple models.\end{minipage}\\
        \addlinespace
        \botrule
    \end{tabularx}
    \label{tab:augsummary}
\end{table}

Within the surveyed approaches, several trends can be observed. Often, the difference in performance between the different transformer architectures (especially \textsc{BERT} and \textsc{RoBERTa} in their varying sizes) is more noticeable than the impact of the proposed augmentations. The actual task being used for benchmarking and the reported performance metrics can also significantly impact how a model's performance may be perceived. This is noticeable in the substantial difference in reported values in Table \ref{tab:augmentations} depending on whether accuracy, F1, or exact match score is used for evaluation.

In addition, ablation studies based on the same base model are not always part of the proposed approaches, making it more difficult to assess the impact of the augmentation itself.

It is also often apparent that the proposed augmentations' performance gains shrink as the base model's performance increases. For example, \textsc{CoCoLM}'s implementation provides a 19.3 percent increase in performance over a base \textsc{BERT} model on debiased \textsc{ROCStories}. However, it improves only 1.3 percent over a \textsc{RoBERTa}-large model. These results indicate that larger models may already possess most of the reasoning capabilities that some of the proposed enhancements can offer.

The prevalence of \textsc{McTaco} compared to other benchmarking datasets is also notable and may be due to their more detailed taxonomy. Approaches that focus on different TCS dimensions can use a subset of \textsc{McTaco} to test performance in the corresponding dimension (e.g., \citet{eventoptimaltransport,sleer}). 

We propose the following future work to enhance TCS reasoning.

\begin{itemize}
    \item Models are generally fine-tuned and tested on only one of the proposed benchmark datasets. Transfer learning could be explored further in several ways, including whether models trained on one dataset perform better on others in a few-shot setting or whether ensembling similar models trained on different TCS reasoning datasets improves overall reasoning capabilities.
    \item A more thorough investigation of the performance of commonsense LMs in downstream tasks would be interesting. For example, \citet{timebert} apply their \textsc{BERT}-based model for document dating as a component in a temporal question answering system, improving overall performance on the downstream task. Possible application areas for the proposed models could be timeline summarization or question answering.
    \item In general, despite increased efforts, the models proposed so far do not reach human performance, even on overly forgiving metrics such as accuracy and F1-score, which do not strongly discredit shallow reasoning compared to metrics like EM or contrast set consistency. Dimensions such as \emph{event typical time}, \emph{stationarity}, and \emph{event frequency} are especially underexplored. New models aiming to reason over these properties could add a new perspective to the overall understanding of TCS and provide new possibilities for downstream applications (such as user status tracking or recommender systems).
\end{itemize}

\subsection{Foundation Models and Trade-offs}
Foundation models, such as \textsc{GPT-4}, have become increasingly influential in recent years. With larger and more rigorously trained models such as \textsc{RoBERTa} outperforming \textsc{BERT}, a logical conclusion might be that the trend towards large foundation models would render much of the previous research obsolete. For several reasons, we do not believe this to be the case.

\begin{itemize}
    \item Current consumer hardware quickly reaches its limits when training and prompting state-of-the-art LMs. Even if the \textsc{GPT-4} weights were publicly known, it is unlikely that most people could run the model locally at a reasonable speed. For privacy reasons, among others, it is therefore unreasonable to expect individuals and businesses to rely fully on API-based prompting.
    \item While foundation models are trained in general language understanding, locally trained models have weights that are specifically fine-tuned on a certain task, making them more of a ``master of one'' than a ``jack of all trades''. For downstream task applications, this is likely preferable.
    \item When training a task-specific machine learning model, we can reliably force the output to be of a certain shape and represent certain task-specific properties. While it is possible to prompt systems like \textsc{GPT-4} to provide a certain output format, they are not constrained to this shape and can deviate, potentially leading to errors.
\end{itemize}

In addition to the arguments provided, recent research shows that foundation models seemingly do not fully comprehend TCS. For example, \citet{chatgpt-tcs} measure \textsc{GPT-3.5} with an accuracy of 53 percent on \textsc{McTaco}, far below previous state-of-the-art work. \citet{llm-eval} benchmark a combination of various foundation models and prompting strategies on several of the datasets discussed in this survey. While they measure \textsc{GPT-3.5} at an accuracy of 80 percent on \textsc{McTaco}, the performance on several other datasets, including \textsc{TNLI}, \textsc{TIMEDIAL}, and \textsc{WikiHow}, remains subpar and lags far behind the baseline model performances reported by the respective dataset authors.

This creates a strong argument that TCS will remain a valuable research field in the near future. The difference in accuracy between the two studies on \textsc{GPT-3.5} also shows the volatility of using closed-source foundation models, as well as different prompting strategies.

\section{Summary and Outlook}
\label{sec:summary}

In this survey, the history of temporal reasoning and the shift to LMs for commonsense reasoning has been explored. Temporal reasoning mainly started as a purely logical proposition over specific data structures, such as Allen's interval algebra. However, as computers became more powerful, syntactic approaches started to emerge. These approaches first centred on manual annotation of events and temporal quantifiers in free-form text, and then gradually moved to how these annotations could be automated using syntactic features such as parse trees and the meaning of specific signal words. 

However, this syntactic analysis lacks the TCS understanding that can be vital in reasoning over free-form text, as time is often only implicitly described in language. This first led to the implementation of semantic features such as semantic role labels and later to a shift to data-driven approaches to solve specific tasks, such as event extraction. The introduction of word embeddings and the subsequent rise of deep neural networks such as LSTMs and transformers allowed for better performance on new tasks and ones derived from previous temporal reasoning propositions. Models trained on these new tasks no longer wholly rely on explicit temporal context, making them useful in domains where such context may not usually be available.

In light of this, it is easy to say that transformers are a plug-and-play solution to TCS tasks, as they outperform previous state-of-the-art methods by a wide margin even when not fine-tuned. However, on closer inspection, it is clear that while the semantic reasoning performed by transformer models is powerful, they are prone to linguistic traps, are not always reliable in their answers, and do not reason over temporal properties as well as we would like. Specifically, mispriming and reporting bias are still significant problems in transformer models when using them for common sense purposes. In addition, they can behave somewhat erratic when the input slightly changes or, conversely, they can ignore critical negations or contrasting data instances.

Several methods have been proposed to overcome this problem. For example, more training data specific to temporal properties can be created via crowdsourcing, rule-based extraction from the web (leading to weak supervision), or from KGs, which themselves can be either manually created (such as \textsc{ATOMIC}$^{20}_{20}$) or constructed automatically (such as \textsc{ASER}). Ensembling of several neural models has also been shown to somewhat increase performance, which is particularly interesting due to the variety of TCS datasets which are now available.

Another intriguing trend is infusing hard-coded symbolic or logical reasoning back into transformer models through methods such as probabilistic soft logic or using the neural model's output as symbolic information to explicitly reason over with traditional methods. This augmentation is especially beneficial in leveraging our innate understanding of language and how temporal traits such as duration, distance, and frequency of events interact. This explicit knowledge often seemingly outperforms attempts to implicitly infuse this understanding into models through auxiliary objectives during training. 

Additionally, the transformer landscape is still evolving, and the importance of aspects such as training objectives, training data, encoding, and hyperparameters such as masking probability cannot be overstated. This can easily be observed by inspecting the difference in accuracy between base \textsc{BERT} and \textsc{RoBERTa} on many tasks. \citet{t5} provide further insight on the importance of such properties. On the other hand, the architecture surrounding the transformer model also impacts performance. With models reaching sizes that make training a state-of-the-art transformer model from scratch difficult or even infeasible for the average person, it is currently much easier to try to improve the architecture surrounding the transformer model rather than the structure of the model itself. 

Ultimately, the goal would be to achieve human performance on the currently proposed and future TCS tasks, allowing for more downstream task applications. To that end, the research community should strive to make transformer-based models more resilient to mispriming and linguistic traps and teach them to better recognize the temporal properties of events and their meaning. To achieve human performance, a model's understanding of explicit and implicit temporal signals must go beyond the individual meaning of each word and co-occurrence statistics learned during pre-training. The significant performance gap between model- and human performance when using more robust metrics signals that models often end up with the correct answer more by clever guesswork, pattern recognition, and biases, but cannot truly reason over words in the same way humans can. Novel methods for infusing logical traits of language and symbolic knowledge into transformer models would likely go a long way to improving this situation.

Overall, the field of TCS reasoning is all but solved, and many avenues for improvement are unexplored so far. Despite preliminary results on \textsc{GPT-4} indicating unprecedented performance on many reasoning tasks, its size and the lack of reporting on the model structure and weights render it unreasonable for local deployment and usage. Additionally, TCS reasoning capabilities may be somewhat limited even in models like \textsc{GPT-4}. Therefore, it is unlikely that foundation models will subsume research into smaller and more portable transformer models.

\backmatter

\section*{Declarations}

\subsection*{Funding}
The authors did not receive support from any organization for the submitted work. 

\noindent No funding was received to assist with the preparation of this manuscript.

\noindent No funding was received for conducting this study. 

\noindent  No funds, grants, or other support was received.

\subsection*{Competing Interests}
The authors have no relevant financial or non-financial interests to disclose.

\noindent The authors have no conflicts of interest to declare that are relevant to the content of this article.

\noindent All authors certify that they have no affiliations with or involvement in any organization or entity with any financial interest or non-financial interest in the subject matter or materials discussed in this manuscript. 

\noindent The authors have no financial or proprietary interests in any material discussed in this article.

\subsection*{Author Contributions}
CRediT author statement:

\noindent \textbf{Georg Wenzel}: Conceptualization, Methodology, Validation, Investigation, Data Curation, Writing - Original Draft \textbf{Adam Jatowt}: Writing - Review \& Editing, Supervision

\bibliography{main}

\end{document}